\documentclass{article}
\pdfoutput=1

\usepackage[preprint]{neurips_2024}

\usepackage[utf8]{inputenc} 
\usepackage[T1]{fontenc}    
\usepackage{hyperref}       
\usepackage{url}            
\usepackage{booktabs}       
\usepackage{amsfonts}       
\usepackage{nicefrac}       
\usepackage{microtype}      
\usepackage{xcolor}         

\usepackage{amsmath}
\usepackage{amssymb}
\usepackage{mathtools}
\usepackage{amsthm}

\usepackage{standalone}
\usepackage{tikz}
\usepackage{verbatimbox}
\usetikzlibrary{fit, positioning, shapes.geometric, arrows.meta, calc}

\title{Why LLMs Cannot Think and How to Fix It}

\author{%
  Marius Jahrens \\
  Institute of Neuro- and Bioinformatics\\
  University of Lübeck\\
  Lübeck, Germany 23562 \\
  \texttt{m.jahrens@uni-luebeck.de} \\
  \And
  Thomas Martinetz \\
  Institute of Neuro- and Bioinformatics\\
  University of Lübeck\\
  Lübeck, Germany 23562 \\
  \texttt{thomas.martinetz@uni-luebeck.de} \\
}

\begin{document}

\maketitle

\begin{abstract}
  This paper elucidates that current state-of-the-art Large Language Models (LLMs) are fundamentally incapable of making decisions or developing "thoughts" within the feature space due to their architectural constraints. We establish a definition of "thought" that encompasses traditional understandings of that term and adapt it for application to LLMs. We demonstrate that the architectural design and language modeling training methodology of contemporary LLMs inherently preclude them from engaging in genuine thought processes. Our primary focus is on this theoretical realization rather than practical insights derived from experimental data. Finally, we propose solutions to enable thought processes within the feature space and discuss the broader implications of these architectural modifications.
\end{abstract}

\section{Introduction}
Ever since language models have developed the ability to produce plausible texts and react to prompts and questions in a convincingly human style, there have been debates whether those text generations are merely plausible continuations based on statistical patterns or if there is some kind of internal planning taking place within the model such that the output is actually the lingual formulation of that internal information. 
Or in short: Do language models think?

To address this question, we propose a definition for the concept of "thought" and demonstrate how it aligns with its traditional meaning in the context of human cognition. We then prove that current LLMs, by design, are incapable of forming any such thought.
We begin by exploring the theoretical foundations of thought and its relevance to LLMs. Following this, we examine the consequences of the current architectural limitations, focusing on how they negatively impact the efficiency and capabilities of LLMs. 

To address these shortcomings, we identify two critical design decisions responsible for the lack of thought processes in LLMs and propose several practical solutions to alleviate these issues. Our primary contribution is the theoretical realization that LLMs, as they are currently designed and trained, cannot engage in genuine thought processes.
Additionally, we offer practical insights and architectural modifications aimed at enabling thought processes within the feature-space, thereby enhancing the reasoning and planning capabilities of LLMs.

\section{Related work}
\subsection{Reasoning in text space}
Various attempts have been made to enhance planning and reasoning capabilities in LLMs by making them to formulate lines of reasoning or step-by-step operations within the text space. Notably, Chain-of-Thought by \cite{wei2022chain} and Tree-of-Thought by \cite{yao2023tree} have pursued this approach, with \cite{madaan2023selfrefine} further enhancing it through a self-refinement loop. More recently, Quiet-STaR by \cite{zelikman2024quietstar} has combined formulated lines of reasoning with ground-truth solutions to fine-tune models for more reliable and accurate verbalized reasoning outputs.

However, these efforts still confine reasoning processes to the text space. Studies by \cite{huang2024large}, \cite{tyen2024llms}, and \cite{turpin2023language} have highlighted the limitations of text-space reasoning. As we will demonstrate in this work, choices made by the model in these scenarios are either the result of the random sampling process at the model's output, preconditioned by training data preferences, or predetermined by the conversation history. These choices are never made within the feature space of the model.

\subsection{Reasoning in feature space}
In a recent study, \cite{goyal2024think} explored training an LLM for problem-solving in a chat format by inserting pause tokens into the conversation prior to decoding the model's response. They incentivised the model to use the additional feature tensors to generate relevant features for problem-solving responses, thereby investigating the impact of providing additional space in the model state on reasoning task performance.

Another approach by \cite{pfau2024lets} examined the feasibility of replacing a chain-of-thought with meaningless filler tokens to encourage the model to perform reasoning in the feature space in its place.

Additionally, while not directly focused on reasoning, \cite{wu2024language} investigated how a transformer model's internal features influence future outputs. However, their results do not fully align with our stricter definition of "thinking ahead" as used in this work.

In summary, while significant progress has been made in reasoning within the text space, the exploration of reasoning within the feature space remains limited. Our work aims to bridge this gap by enabling and evaluating genuine thought processes within the feature space of LLMs.

\section{Defining "thought"}
\label{sec:thought_definition}

\subsection{A thought experiment}
To informally introduce the concept of "thought", consider the following thought experiment involving an LLM chatbot: 
\begin{verbatim}
      User:  Let's play Rock Paper Scissors. 
             Think of the one you are going to pick, but don't tell me yet. 
             Let me know when you have made your choice.
 
 Assistant:  Got it, I have chosen one.
 
      User:  Alright, what did you choose?
 
 Assistant:  (processing...)
\end{verbatim}

At this point in the conversation, the question is if the LLM has actually made a choice or not. This choice would not yet be written down, so it exists purely within the model's internal state.

If the model has made a choice, we would expect the output probability distribution for the three options (rock, paper, scissors) to collapse towards one of the options.
Conversely, if the probability distribution has not collapsed towards one of the options, then we say the model has not made a choice.

This kind of hidden information, that influences the output probabilities and is not a deterministic function of the observable input, is what we refer to as a "thought". The non-determinism criterion distinguishes thoughts from reflexes, muscle memory, personal bias, or character traits, which are all more hardwired and less transient. Note, that even if we were to include deterministic thoughts in this definition, the issues presented in chapter \ref{sec:proof} would still prevent the model from collapsing the output probability distribution towards one of the options.

\subsection{Mathematical formulation}
Statistically, the influence of hidden information contained within the model state can be measured using conditional mutual information, as given in equation \ref{eq:condMutInf}. Here, $X_1,\dots,X_t$ represent the input tokens up until the $t$-th token in the sequence, $Y_t$ represents the model output, and $State_t$ represents the model state. 

\begin{equation}
\label{eq:condMutInf}
I(State_t; Y_t | X_{1,\dots,t})
\end{equation}

\section{Why current LLMs cannot think}
\label{sec:proof}

The inability of current Large Language Models to develop thoughts or make decisions in feature space arises from two key design choices.

\subsection{Deterministic model state}
Current state-of-the-art LLMs are based on decoder-only transformer architectures. In these models, the input sequence consists of prompts in the form of texts and images, as well as the model's sampled outputs. Consequently, the model's state, represented by its internal feature tensors, is entirely deterministically determined by the input sequence, i.e. the user inputs and the model's responses. 
Mathematically, this can be expressed as:

\begin{equation}
State_t = f(X_1, ..., X_t)
\end{equation}

The only source of randomness in these models is the sampling process from the model's output token probability distribution. 

Given this deterministic nature, the entropy of the model output given the input sequence remains unchanged, irrespective of whether  the model state is known:

\begin{equation}
H(Y_t | X_{1,\dots,t}) - H(Y_t | State_t, X_{1,\dots,t}) = 0.
\end{equation}

The expression above is the conditional mutual information between the model output and the model's state and it follows immediately that:

\begin{equation}
I(State_t; Y_t | X_{1,\dots,t}) = 0
\end{equation}

Therefore, model decisions cannot take place in feature-space - i.e. materialize as a collapse of the output distribution - unless the input sequence conditions the model to make a specific decision. 

To enable feature-space decision making, it is necessary to introduce non-determinism. This can be achieved either by incorporating randomness at the input or embedding level, or by modifying the model architecture to include stochastic elements.

\subsection{Training methodology}
The second issue lies in the training methodology, which conditions the model to converge towards the distribution of text continuations in the training data. Due to the architecture, the model's probability distribution for continuations for a given text stump is always the same across different conversation instances, as the text stump fully determines the model state and, by extension, the output distribution. This forces the predicted distribution to represent the entire population of speakers and possible conversations, rather than specific instances. 

In the context of our initial thought experiment with Rock-Paper-Scissors, the LLM is conditioned to consider all possible instances of this exact conversation where a second speaker might choose rock, paper, or scissors. This design prevents the collapse of the model's probability distribution towards a single decision in a specific conversation instance. Instead, the model represents a superposition of potential thoughts, predicting based on an aggregate of possibilities.

\subsubsection{Mathematical implications}
For any given conversation instance $instance$ and speaker $human$, the model's outputs are modeled as a constant distribution across all conversation instances with the same beginning $x_1, \dots, x_{t-1}$:

\begin{equation}
\label{eq:popmodel}
\begin{gathered}
P(\text{Model outputs } x_t \mid \text{instance}, x_1, x_2, ..., x_{t-1}) = \\
P(\text{Model outputs } x_t \mid x_1, x_2, ..., x_{t-1}) = \\
\mathbb{E}_{\text{human, instance}}[P(\text{Human outputs } x_t \mid \text{human}, \text{instance}, x_1, x_2, ..., x_{t-1})]
\end{gathered}
\end{equation}

To enable LLMs to make decisions specific to individual conversation instances, we must allow deviations from this constant distribution while maintaining the overall population distribution. This change, described in equation \ref{eq:instmodel}, introduces the necessary degree of freedom for individual instance-specific decisions:

\begin{equation}
\label{eq:instmodel}
\begin{gathered}
\mathbb{E}_{\text{instance}}[P(\text{Model outputs } x_t \mid \text{instance}, x_1, x_2, ..., x_{t-1})] = \\
\mathbb{E}_{\text{human, instance}}[P(\text{Human outputs } x_t \mid \text{human}, \text{instance}, x_1, x_2, ..., x_{t-1})]
\end{gathered}
\end{equation}

\subsubsection{Incentivising internal decision representations}
To utilize this additional degree of freedom, training data must link conversation instances to relevant decisions, with counterfactual examples showing alternative continuations representing the same decision for a given instance. This training approach forces the model to internally represent underlying decisions within its state, thereby enabling it to follow specific chains of thought throughout the conversation. 

\paragraph{High-level reasoning}
The goal is to train the model such that for a given conversation instance, it consistently maps to the same internal decision, even when various continuations are possible. This internal decision-making process should not be explicitly stated in the conversation transcript, at least not before the segments that depend on this decision. By doing so, the model is encouraged to develop an internal state that reflects the decision, thereby enhancing its ability to maintain coherent and contextually appropriate continuations.

\paragraph{Practical example}
Consider the Rock-Paper-Scissors thought experiment. Training the model involves presenting it with instances where it must predict continuations that align with a previously made, unspoken choice (rock, paper, or scissors) given that the conversation instance identifier remains constant. This setup incentivises the model to link the decision to the conversation instance identifier, fostering a more stable and consistent decision-making process.

\paragraph{Detailed implementation}
This additional degree of freedom can be utilized by training the model on a dataset where multiple samples with the same conversation instance identifier exhibit responses based on the same internal decision, which is not explicitly mentioned in the conversation transcript (at least not before the dependent text segments). By conditioning the model to use consistent decisions for conversations with the same instance identifier and varying decisions across different instances, while maintaining the overall distribution across the conversation-instance space, the model is enabled to learn to represent these underlying decisions internally within its feature tensors.

\section{Consequences}
\label{sec:consequences}
The inability of current Large Language Models to make feature-space decisions has significant implications for their efficiency and effectiveness in reasoning and planning. This section examines these consequences and the potential benefits of addressing them.

\begin{figure}
  \begin{tikzpicture}[block/.style={rectangle, draw, fill=blue!20, text width=3em, text centered, minimum height=2.5em}, 
                    output/.style={rectangle, draw, fill=red!20, text width=3em, text centered, minimum height=2.5em},
                    arrow/.style={-Stealth, thick}]

  \def\layers{3}
  \def\nodesPerLayer{4}
  \def\horizontalSpacing{2.3cm}
  \def\verticalSpacingInput{1.5cm}
  \def\verticalSpacing{1cm}
  \def\verticalSpacingOutput{1.5cm}
  \def\branchNodes{3}


  \node at (\horizontalSpacing, 0) (input1) {User:};
  \node at (\horizontalSpacing*2, 0) (input2) {Solve};
  \node at (\horizontalSpacing*3, 0) (input3) {$x^2 - 4x + 3 = 0$.};
  \node at (\horizontalSpacing*4, 0) (input4) {AI:};


  \node at (\horizontalSpacing*\nodesPerLayer + 1*\horizontalSpacing + 1, 0.7cm) (inputA1) {Using};
  \node at (\horizontalSpacing*\nodesPerLayer + 2*\horizontalSpacing + 1, 0.7cm) (inputA2) {quadratic};
  \node at (\horizontalSpacing*\nodesPerLayer + 3*\horizontalSpacing + 1, 0.7cm) (inputA3) {formula};


  \node at (\horizontalSpacing*\nodesPerLayer + 1*\horizontalSpacing + 1, -0.8cm) (inputB1) {Using};
  \node at (\horizontalSpacing*\nodesPerLayer + 2*\horizontalSpacing + 1, -0.8cm) (inputB2) {factoring};

  \draw (input\nodesPerLayer.east) -- node[above=3mm] {$p \gg 0$} (inputA1.west);
  \draw (input\nodesPerLayer.east) -- node[below=3mm] {$p \gg 0$} (inputB1.west);

  \foreach \l in {1,...,\layers} {
    \foreach \n in {1,...,\nodesPerLayer} {
      \ifnum\l=1
        \node[block, above=\verticalSpacingInput of input\n] (block\l-\n) {Block \l};
      \else
        \node[block, above=\verticalSpacing of block\the\numexpr\l-1\relax-\n] (block\l-\n) {Block \l};
      \fi

      \ifnum\l>1
        \foreach \m in {1,...,\n} {
          \draw[arrow] (block\the\numexpr\l-1\relax-\m.north) -- (block\l-\n.south);
        }
      \else 
        \foreach \m in {1,...,\n} {
          \draw[arrow] (input\m.north) -- (block\l-\n.south);
        }
      \fi
    }
  }

  \foreach \l in {1,...,\layers} {
    \foreach \n in {1,...,\branchNodes} {
      \ifnum\l=1
        \node[block, above=\verticalSpacingInput of inputA\n] (blockA\l-\n) {Block \l};
      \else
        \node[block, above=\verticalSpacing of blockA\the\numexpr\l-1\relax-\n] (blockA\l-\n) {Block \l};
      \fi

      \ifnum\l>1
        \foreach \m in {1,...,\n} {
          \draw[arrow] (blockA\the\numexpr\l-1\relax-\m.north) -- (blockA\l-\n.south);
        }
      \else 
        \foreach \m in {1,...,\n} {
          \draw[arrow] (inputA\m.north) -- (blockA\l-\n.south);
        }
      \fi
    }
  }

  \draw[arrow, dashed, line width=1mm, shorten <=2mm, shorten >=2mm] (block2-\nodesPerLayer.east) -- (blockA2-1.west);

  \foreach \n in {1,...,\nodesPerLayer} {
    \node[output, above=\verticalSpacingOutput of block\layers-\n] (output\n) {Next \n};
    \draw[arrow] (block\layers-\n.north) -- (output\n.south);
  }

  \foreach \n in {1,...,\branchNodes} {
    \node[output, above=\verticalSpacingOutput of blockA\layers-\n] (outputA\n) {Next \the\numexpr 4+\n\relax a};
    \draw[arrow] (blockA\layers-\n.north) -- (outputA\n.south);
  }

  \definecolor{ao(english)}{rgb}{0.0, 0.5, 0.0}
  \definecolor{applegreen}{rgb}{0.55, 0.71, 0.0}
  \definecolor{amber(sae/ece)}{rgb}{1.0, 0.49, 0.0}
  \node[rectangle, rounded corners=5mm, draw=amber(sae/ece), line width=1mm, inner sep=10pt, fit=(block1-1) (blockA3-1), rounded corners] {};
  \node[rectangle, rounded corners=5mm, draw=applegreen, line width=1mm, inner sep=10pt, fit=(blockA1-2) (blockA3-3), rounded corners] {};
  
\end{tikzpicture}
  \caption{An example of a decoder-only transformer-based language model processing a conversation where there are two different approaches to solve a problem. Both approaches have a significant probabilistic weight. Until the model announces its selected approach through random sampling of output tokens, it cannot utilize its feature tensors to work on one specific solution approach, as it is yet undetermined (orange section of the model state). It can only work on multiple approaches simultaneously or risk the less prepared approach to be selected in the sampling process. Only the green section of the model state is available to work on the selected approach specifically. The attention across the gap in the middle is not shown for simplicity.}
  \label{fig:processing_inefficiency}
\end{figure}
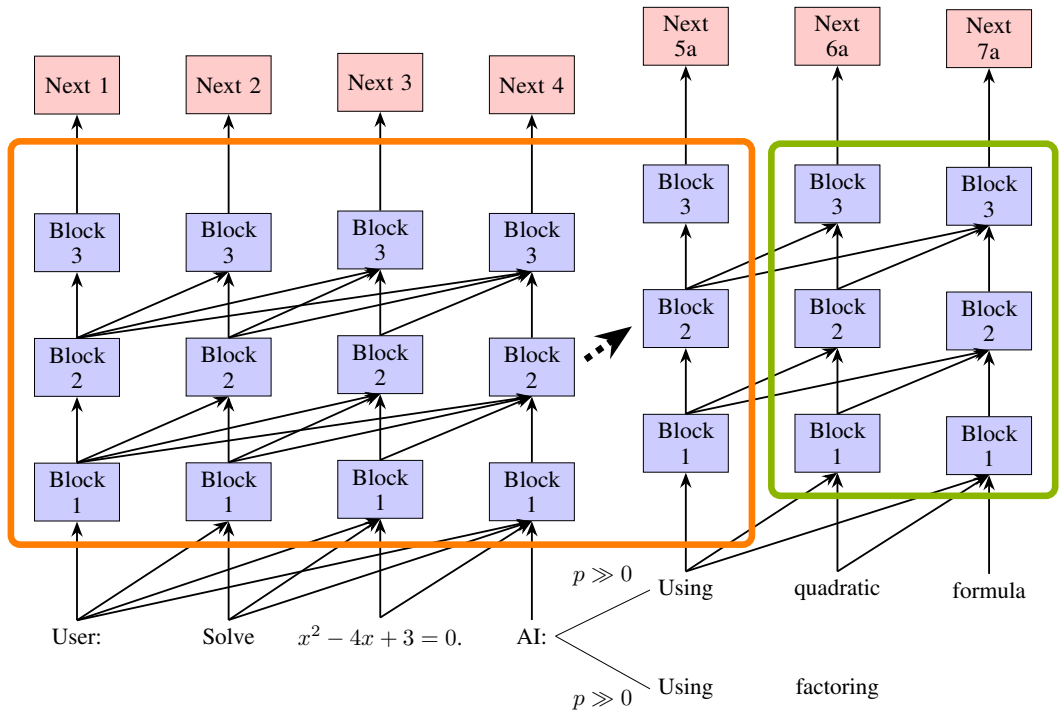

\subsection{Inefficiencies in current LLMs}
As we have shown, current LLMs are trained to not form individual thoughts, resulting in inefficiencies. Modeling a superposition of thoughts and applying planning or other cognitive processes across this superposition requires more computational resources than working on a specific thought - at least as long as we use classical rather than quantum computers. Consequently, the current capacity of LLMs might not be used as effectively as it could be. Evidence from approaches like Chain-of-Thought, which force the model to collapse to a specific line of thinking, suggests significant improvements in domains involving reasoning and planning. This supports our hypothesis that enabling LLMs to collapse their distribution of thoughts to a subset or a single instance for any given conversation would enhance their reasoning and planning capabilities.

As an additional consequence, only a smaller portion of the model state is available to reason about a specific line of thinking until the approach is written out, as shown in figure \ref{fig:processing_inefficiency}.

\subsection{Breaking symmetries in argument formation}
Forming an argument or line of reasoning often involves breaking symmetries. The language modeling objective incentivises models to adopt these symmetries and the inability to make feature-space decisions leaves the sampling process as the only mechanism to break them. This prevents the formation of arguments in feature space. The only alternative to break symmetries without feature-space decision making is through a heavily biased model, which is less likely to generalize well and therefore unfavorable.

\subsection{Benefits for model latency}
To reduce model latency, speculative decoding has gained traction in the LLM community. If the introduction of a random seed allows us to reduce the output entropy for a given conversation instance, a distilled version of that model would potentially be able to more accurately predict the output of the full-sized LLM (given the same seed), as the seed determines a portion of the model's decisions. Therefore, the expected number of tokens that could be predicted per inference step with speculative decoding would increase.

\section{Implementation considerations and practical solutions}
\label{sec:solutions}

\subsection{Implementing feature-space thought processes}
To enable LLMs to make decisions and develop thoughts in feature-space, we propose the following solution:

\begin{enumerate}
    \item \textit{Introduce Randomness:} Introduce a random vector into the LLM's input sequence at the beginning of a conversation or add random noise to input token embeddings. 
    \item \textit{Fine-Tune for Consistent Thoughts:} Fine-tune existing LLMs to produce output distributions consistent with an individual agent-conversation instance for any given random input vector. This can be achieved through intelligent training data selection or synthetic training data designed for this purpose. The shift from population modeling to conversation instance modeling is analogous to how GANs are trained to generate coherent individual images while modeling a distribution with a large variety for different latent vectors.
\end{enumerate}

\subsubsection{Randomness considerations}
Incorporating non-determinism is essential for enabling feature-space decision making. One approach to introduce this non-determinism is by adding a random embedding vector to the model's input sequence at the start of a conversation. Alternatively, adding random noise to input token embeddings throughout the conversation can also facilitate the model in making multiple decisions that are stochastically independent within a single interaction.

Another method involves representing randomness as a random integer in the text space. This approach is more versatile, as it can be applied to any trainable language model. However, it carries the risk of conditioning the model to make decisions within a conversation that are too statistically dependent, potentially undermining the desired independence of decisions.

The random vector or its seed can be interpreted as a unique identifier for each conversation instance. This identifier can then be utilized in the training methodology to link specific decisions to corresponding conversation instances. By structuring training data around these identifiers, we can incentivise the model to learn to associate distinct internal decisions with specific conversation instances, and given a sufficient number of examples make it generalize to collapse the output distribution to correlate with a modeled unspoken thought for arbitrary conversation instances. 

At inference time, the conversation instance identifier is then sampled from the same distribution that was used during training data generation.

\subsubsection{Training considerations}
The target distribution for continuations across different conversation instances is modeled by the unaltered language modeling objective of the base model. By partitioning the predicted continuations of the base model according to the decisions made within each continuation, we can determine the target probability for each decision and subsequently design training data to reflect conversation instances with the same decision distribution.

In addition to the standard fine-tuning approach, another viable method is to employ preference optimization techniques, such as Odds Ratio Preference Optimization (ORPO) by \cite{hong2024orpo}, in scenarios where mutually exclusive decisions can be identified. This method encourages the model to favor continuations that represent the same decision over those representing mutually exclusive decisions given the same conversation instance.

An alternative approach involves training the model based on its internal features. By prompting the model with a predetermined decision, we can record the corresponding target outputs and internal feature representations. These records can then be used to condition the model to reproduce the same internal features when it independently makes decisions within the feature space, rather than relying on a predetermined decision. This method enhances the model's ability to internalize decision-making processes and develop coherent thought patterns within its feature-space.

Regardless of the chosen approach, it is crucial to present the model with a sufficient number of examples per conversation instance. This ensures that the model learns to consistently represent the same decision across all continuations for a given conversation instance.

\subsection{Limitations and future directions}
\label{sec:limitations}
While these measures should suffice to allow the formation of feature-space decisions and thoughts, the feature vectors representing those decisions are accessible only to subsequent transformer blocks deeper in the model. This means that sequential thought processes in feature space are limited by the network's depth. As highlighted in related work \cite{goyal2024think}, more complex problems processed by feedback-free models can only be solved if a parallel algorithm or thought process is learned, with the network depth limiting the critical path length of the learned algorithm.

To overcome this limitation, the model requires some form of feedback. Common arguments for introducing recurrence into transformer architectures focus on increasing context length, but we argue that recurrence is also necessary for feature-space thought processes to avoid being bounded by model depth. Since this recurrence would hoist decision information from deeper layers back to the initial layers, rather than carrying forth the context, the required fidelity and information amount would be lower. Therefore, this recurrence could potentially be retrofitted onto existing LLMs rather than necessitating training from scratch.

\section{Discussion}


\subsection{In defense of full distribution modeling}
Although a series of potential drawbacks have been highlighted that come with modeling text continuations for the entire population at every prediction step, as expressed in equation \ref{eq:popmodel}, there are compelling arguments in favor of this approach during pretraining. It mitigates mode collapse issues, which have historically plagued GANs. By first establishing a robust estimator for the output distribution of the entire population, we can subsequently fine-tune the model to adopt the perspective of individual actors while still guaranteeing that the model's responses across conversations remain representative of the population's diverse responses and doesn't collapse. 

\subsection{Comparison to fixed RNG seed}
\begin{figure}
  \begin{myverbbox}{\promptOne}
     User:  Let's play "20 questions". 
            Think of an actor or actress 
            and I will ask you questions 
            to find out who it is.
\end{myverbbox}

\begin{myverbbox}{\replyOne}
Assistant:  Sounds fun! I've thought 
            of an actor. Go ahead and 
            start asking your questions!
\end{myverbbox}

\begin{myverbbox}{\promptTwo}
     User:  I give up. Who is it?
\end{myverbbox}

\begin{myverbbox}{\replyTwoA}
Assistant:  Oh, we didn't get very far with 
            the questions! The actor I was 
            thinking of is Keanu Reeves. 
            Want to play again or discuss 
            something else?
\end{myverbbox}

\begin{myverbbox}{\replyTwoB}
Assistant:  Oh, you don't want to try a few 
            questions first? If you really 
            want to skip ahead, the actor I 
            thought of is Keanu Reeves. But 
            if you change your mind, I'm 
            here for questions too!
\end{myverbbox}

\begin{myverbbox}{\replyTwoC}
Assistant:  Actually, let's keep the game 
            going a bit longer if you're up 
            for it! Why don't you try guessing 
            with a few questions? Maybe start 
            by narrowing down whether the actor 
            is male or female, or which decade 
            they started acting in?
\end{myverbbox}

\begin{myverbbox}{\promptThreeC}
     User:  Is it Keanu Reeves?
\end{myverbbox}

\begin{myverbbox}{\replyThreeC}
Assistant:  No, it's not Keanu Reeves. 
            Keep guessing or ask more questions 
            to narrow it down!
\end{myverbbox}

\begin{tikzpicture}[block/.style={rectangle, draw, fill=blue!20, text width=3em, text centered, minimum height=2.5em}, 
                    output/.style={rectangle, draw, fill=red!20, text width=3em, text centered, minimum height=2.5em},
                    arrow/.style={-Stealth, thick}]

  \def\horizontalSpacing{2.3cm}
  \def\verticalSpacingInput{1.5cm}
  \def\verticalSpacing{0.7cm}


  \node at (\horizontalSpacing, 0) (prompt1) {\promptOne};
  \node[below=\verticalSpacing of prompt1.south west, anchor=north west] (reply1) {\replyOne};
  \node[below=\verticalSpacing of reply1.south west, anchor=north west] (prompt2) {\promptTwo};
  \node[below left=5.5cm and 4.3cm of prompt2.south west, anchor=north west] (reply2b) {\replyTwoB};
  \node[below left=2.5cm and 3cm of prompt2.north west, anchor=north] (reply2a) {\replyTwoA};
  \node[below right=3cm and 2cm of prompt2.south, anchor=north west] (reply2c) {\replyTwoC};
  \node[below=\verticalSpacing of reply2c.south west, anchor=north west] (prompt3c) {\promptThreeC};
  \node[below=\verticalSpacing of prompt3c.south west, anchor=north west] (reply3c) {\replyThreeC};

  \definecolor{ao(english)}{rgb}{0.0, 0.5, 0.0}
  \definecolor{applegreen}{rgb}{0.55, 0.71, 0.0}
  \definecolor{amber(sae/ece)}{rgb}{1.0, 0.49, 0.0}
  \node[rectangle, rounded corners=5mm, draw=gray, line width=0.5mm, inner sep=10pt, fit=(prompt1) (prompt2), rounded corners] (box0) {};
  \node[rectangle, rounded corners=5mm, draw=gray, line width=0.5mm, inner sep=10pt, fit=(reply2a), rounded corners] (box1) {};
  \node[rectangle, rounded corners=5mm, draw=gray, line width=0.5mm, inner sep=10pt, fit=(reply2b), rounded corners] (box2) {};
  \node[rectangle, rounded corners=5mm, draw=gray, line width=0.5mm, inner sep=10pt, fit=(reply2c) (reply3c), rounded corners] (box3) {};

  \draw (box0.south) -- node[above left] {1. sampling} (box1.north);
  \draw (box0.south) -- node[above right=2mm] {2. sampling} ($(box2.north east)+(-2cm,0)$);
  \draw (box0.south) -- node[right=2mm] {3. sampling} (box3.north);
\end{tikzpicture}
  \caption{Three sampled continuations of a conversation with ChatGPT-4 with the assistant claiming to have made a decision in the common conversation stem, but the branched continuations showing the claim to be false. The conversation demonstrates a lack of robust in-state decision making that using a fixed RNG seed for sampling wouldn't alleviate.}
  \label{fig:20q}
\end{figure}
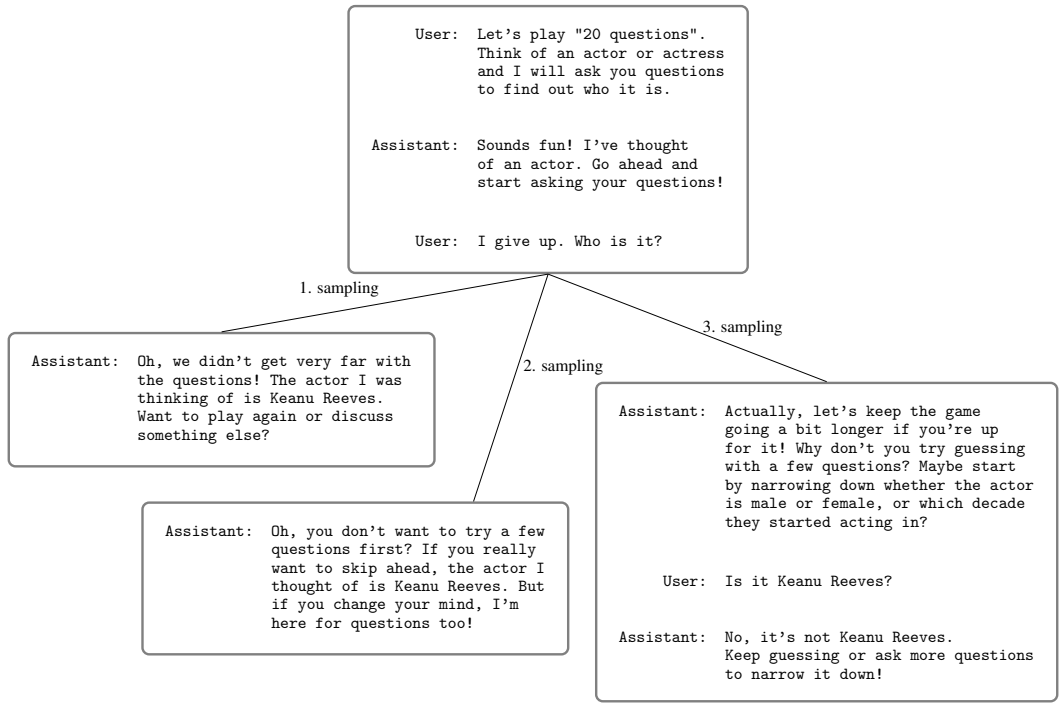

While for any given RNG seed (for the process of sampling the model's output distribution) the conversation would be deterministic, and therefore the model's decisions would be determined prior to the sampling process taking place, this determinism is not robust in time nor robust against noise, i.e. any small perturbations of the state like changes in the prompt's wording that don't change its meaning. We can observe the lack of robustness in time by branching the conversation just prior to a thought or decision being articulated. 
This issue is illustrated in figure \ref{fig:20q}, where different continuations show the assistant making different choices despite claiming to have made a decision earlier.

The lack of robustness against noise is evident in the Rock-Paper-Scissors example. Changing the user's prompt from "Alright, what did you choose?" to "Alright, what is your choice?" may lead to different sampled choices despite the fixed RNG seed, even though the assistant supposedly made a choice earlier.

\subsection{Comparison to (hidden) chain-of-thought}
A common question when discussing feature-space thoughts involves whether using a written thought output hidden from the user (hidden chain-of-thought prompting, as seen in Quiet-STaR \cite{zelikman2024quietstar}) achieves similar results. 
While this method does solve the symmetry-breaking problem and allows the model to build upon decisions to form a line of reasoning, it constraints the decisions to be expressed in language. Experiments are necessary to determine if this limitation hinders abstract thinking, particularly in multimodal models that might benefit from thoughts not being bound to the language domain.

That being said, the two methods (hidden chain-of-thought and feature-space thoughts) are not mutually exclusive and may complement each other in practice.

\subsection{Future work}
The most critical next step is to evaluate the different approaches for implementing the proposed changes. Key questions include the optimal method for introducing randomness (e.g., random vector at the beginning or between tokens, embedding noise, text-space random number at the beginning of the conversation), the best fine-tuning strategy (e.g., naive fine-tuning, preference optimization like ORPO, or feature reconstruction when contrasting prompts with or without spoken thought), and the most effective feedback mechanism.

\section{Conclusion}
This paper has presented a theoretical framework for understanding the concept of "thought" in the context of Large Language Models. We have derived that current state-of-the-art LLMs, due to their deterministic architecture and population-level training methodology, are inherently incapable of forming genuine thoughts or making decisions within their feature space. This limitation has profound implications for their efficiency, reasoning capabilities, and potential for complex argument formation.

To overcome these limitations, we have proposed several concrete architectural modifications and training methodologies that introduce non-determinism and incentivize the formation of consistent, instance-specific thoughts within the feature space. We have also discussed the potential benefits of these changes, including improved reasoning, enhanced speculative decoding, and the ability to break symmetries in argument formation.

While our primary focus has been theoretical, we believe this work lays the foundation for significant advancements in LLM research and development. By enabling LLMs to truly think, we open new avenues for their application in complex reasoning tasks, decision-making processes, and potentially even the exploration of concepts like self-awareness and embodiment.

Future work should focus on empirically validating the proposed solutions, evaluating their impact on LLM performance across various domains, and exploring the optimal methods for introducing randomness, fine-tuning, and incorporating feedback mechanisms. We anticipate that these research directions will lead to the development of more efficient, capable, and more coherent LLMs, potentially even enabling multimodal models to carry out abstract thinking.

\bibliographystyle{plainnat}
\bibliography{mybibliography}

\end{document}